\documentclass[conference]{IEEEtran}
\usepackage{cite}
\usepackage{amsmath,amssymb,amsfonts}
\usepackage{algorithmic}
\usepackage{graphicx}
\usepackage{placeins}
\usepackage{textcomp}
\usepackage{xcolor}
\usepackage{todonotes}
\usepackage{etoolbox}
\usepackage{fancyhdr}
\def\BibTeX{{\rm B\kern-.05em{\sc i\kern-.025em b}\kern-.08em
    T\kern-.1667em\lower.7ex\hbox{E}\kern-.125emX}}
    
\fancypagestyle{IEEEtitlepagestyle}
{
 
   \fancyhf{}
\cfoot{Copyright Aret\'e Associates, Inc. 2021}
}

\fancypagestyle{plain}
{
 
   \fancyhf{}
\cfoot{Copyright Aret\'e Associates, Inc. 2021}
}

\begin{document}
\title{Matching Targets Across Domains with RADON, the Re-Identification Across Domain Network}

\author{\IEEEauthorblockN{Cassandra Burgess}
\IEEEauthorblockA{\textit{Aret\'e }\\
cburgess@arete.com}
\and
\IEEEauthorblockN{Cordelia Neisinger}
\IEEEauthorblockA{\textit{Aret\'e }\\
cneisinger@arete.com }
\and
\IEEEauthorblockN{Rafael Dinner}
\IEEEauthorblockA{\textit{Aret\'e }\\
rdinner@arete.com }}

\maketitle

\begin{abstract}
We present a novel convolutional neural network (CNN) that learns to match images of an object taken from different viewpoints or by different optical sensors. Our Re-Identification Across Domain Network (RADON) scores pairs of input images from different domains on similarity.  Our approach extends previous work on Siamese networks and modifies them to more challenging use cases, including low- and no-shot learning, in which few images of a specific target are available for training.  RADON shows strong performance on cross-view vehicle matching and cross-domain person identification in a no-shot learning environment. 

\end{abstract}

\section{Introduction}
The proliferation of imaging sensors, from satellites to smartphones to traffic cameras, presents an unprecedented opportunity to automatically identify and maintain custody of an object of interest. A great deal of work in deep learning and image processing has been done to improve the detection and identification of objects in imagery. However, with imagery from so many different domains - multi-spectral, RGB, infrared, different viewing angles, etc. - comes the challenge of identifying the same target in different types of images. To address this challenge, we introduce the Re-Identification Across Domain Network (RADON), a convolutional neural network (CNN) that aims to identify matching targets from different sources. 

RADON is useful for problems in which multiple sensors are available and we need to identify the same target across these sensors. In addition, while copious datasets exist for the most common image domains, such as RGB, labeled data for rarer domains are often difficult to obtain, particularly for infrequently seen targets. By transferring knowledge of a target from a common domain to a new domain, RADON can assist in building up new, automatically labeled datasets, enabling research in a broader array of image domains without requiring further human labeling efforts.

Previous deep learning work has been successful in localizing and characterizing targets in imagery given training and testing data from the same domain \cite{alien}. Similarly, networks that have been successful at image matching tasks, such as facial recognition \cite{song2019occlusion} or target re-identification \cite{tracker}, are typically applied within a single domain with both input images coming from the same type of sensor. With RADON, we expand the application of CNNs to cross-domain problems, such as matching faces between RGB and infrared (IR) imagery.

We present the architecture for RADON, a modified siamese network using pre-trained weights, and initial results for two applications: Cross-view matching of vehicles in RGB imagery, and person re-identification between RGB and infrared imagery, including no-shot learning on novel target classes. We achieve our state-of-the-art performance through judicious use of pre-trained weights and modifications to the standard training scheme.


\section{Background} 
\subsection{Siamese Networks} 
\begin{figure}
\includegraphics[width=\columnwidth]{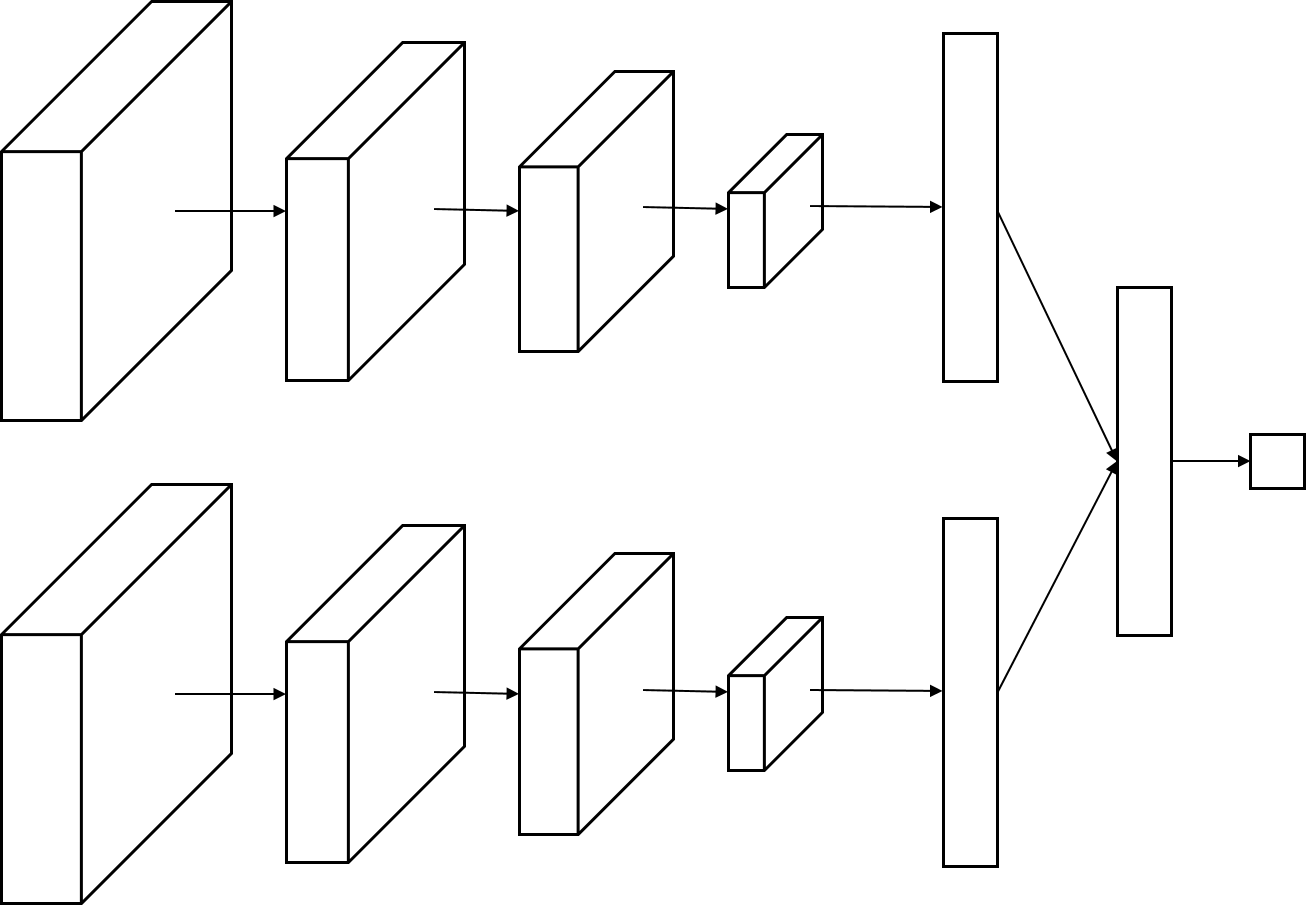}
\caption{A Siamese Network consists of two branches of convolutional layers, combined at the end to produce a single output.} 
\label{fig:siam_arch}
\end{figure}

Broadly, siamese networks are a class of deep neural network composed of two branches that merge together in the final layers to form a single output. Figure~\ref{fig:siam_arch} shows the general architecture for a Siamese network. 

Each branch consists of a series of convolutional layers, which reduce the input images to a feature vector. While each branch processes its input image independently, in a standard siamese network, the weights of the two branches are shared. During training the weights of each branch are updated identically, meaning that the second branch will essentially be a copy of the first. Both input images are thus processed in the same way.

The two branches are merged together by comparing their resulting feature vectors using a distance metric. Additional layers can be added to the network following the merged layer. 

\subsection{Transfer Learning} 
Transfer learning is a method of network training in which we use weights from a network trained on a different problem or dataset to initialize our network, and then update or fine-tune some or all of those weights for the new problem. Previous work has shown great success using transfer learning \cite{transferlearn}. Because labeled datasets for cross-domain matching applications are rare and expensive to generate, we take a transfer learning approach to train our network. This enables us to train a robust, high-performing network without requiring large amounts of data. 

In transfer learning we transfer information from another network and dataset to our own. Typically, transfer learning relies on using weights learned from a problem with large amounts of data available. One such dataset is ImageNet \cite{imagenet}. ImageNet contains over 14 million labeled RGB images for training image classification tasks, enabling networks to learn broadly applicable image features. 

VGG-16 is a frequently used network in transfer learning applications. The network is a very deep convolutional network with 16 convolutional layers, described in \cite{vgg16}. The network with weights trained on ImageNet is available as part of the Keras library in python \cite{keras}. 


\section{Network Approach} 
\subsection{Overview} 
RADON is a variation on a Siamese network with two branches, one for each domain of interest. Each branch takes in an input image, passes through a series of convolutional layers, and produces a final feature vector. The distance between these vectors is computed, and used to determine if the two input images contain the same target. RADON does not tie the weights between the two branches, but rather updates them independently, although they begin with the same weights. This approach allows the network to learn different features for each domain, while taking advantage of transfer learned information for both. Note that we assume that small target chips have already been extracted by an upstream algorithm, and that the target is centered in the input image. 

\subsection{Network Architecture}
\begin{figure}[h]
\includegraphics[width=\columnwidth]{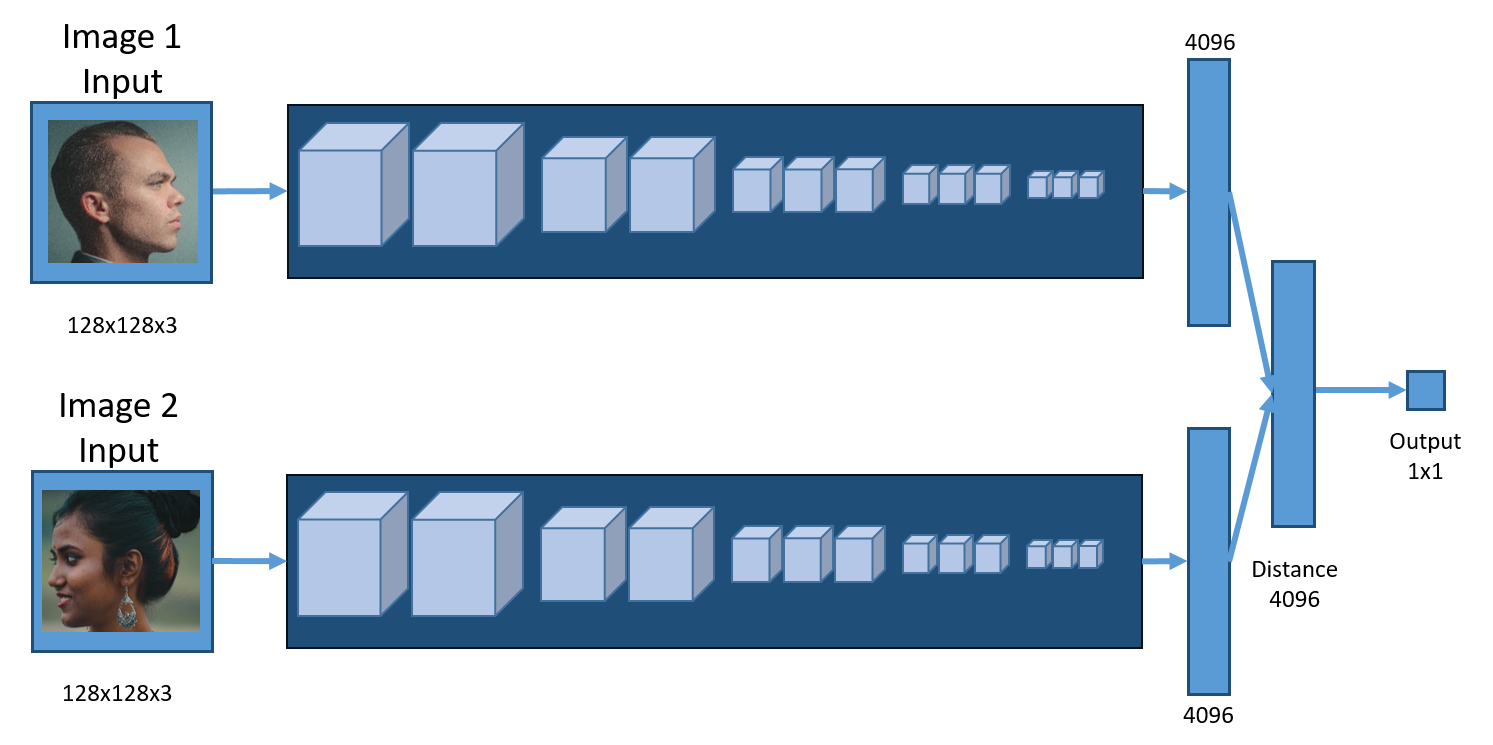}
\centering
\caption{The RADON architecture consists of a
2-branch Siamese Network that takes in two inputs images and
outputs a binary prediction of match/no match. RADON has
previously shown success on several other cross-domain
applications.}
\label{fig:radon_arch}
\end{figure}

RADON takes in two images and outputs a probability
that the pair consists of matching subjects.To accomplish this, RADON uses a modified Siamese Network architecture,
as shown in Figure~\ref{fig:radon_arch}.

Each branch of RADON has a series of convolutional layers that
extract features from the input image. In contrast to a standard Siamese network, the branches in RADON are not tied together. Their weights are updated separately during training, allowing them to specialize to each input domain. At
the end of each branch a final feature vector is generated.
These features are then compared using the L1-distance, 
and the difference is used to
predict whether the input images are of the same target or not.

RADON builds on previous work on convolutional neural networks
for images by incorporating pre-trained weights from similar
problems into the branches. Using a pre-trained generic network,
such as VGG-16 on ImageNet, provides basic image analysis
features to the network, and allows it to successfully train on
a new problem with less time and data. We found that using these pre-trained weights improved performance even when the input image is from a different input domain. 

\subsection{Training}
During training, pairs of input images are passed as input, with a corresponding truth of 1 if the same target is present and 0 otherwise. We provide a balanced training set with 50\% matching images and 50\% non-matching. 

The network was trained with a simple binary cross-entropy loss function and a standard Adam optimizer. Most interestingly, we found no benefit to freezing any of the pre-trained weights at any point during training. 


\section{Application to Vehicle Recognition} 
Using Oak Ridge National Laboratory’s Overhead Vehicle Dataset (OOVD) \cite{oovd}, we tested RADON’s performance on cross-view matching of vehicle targets. In this instance, we match targets across extremely different view angles, where each domain is one angle. Both domains are captured in RGB images. OOVD provides labeled side and overhead images of vehicles on roads at ORNL. We split this data into 616 classes for training and validation and 67 hold-out classes for no-shot testing. No-shot testing or no-shot learning is a term used to describe the process of training a network in which no examples of some class of object are presented at training time.  This is typically used to discover novel targets or anomalies. Here, we ask the network to match image pairs from such unseen classes at testing time.

\begin{figure}[h]
\centering
\includegraphics[width=0.9\columnwidth]{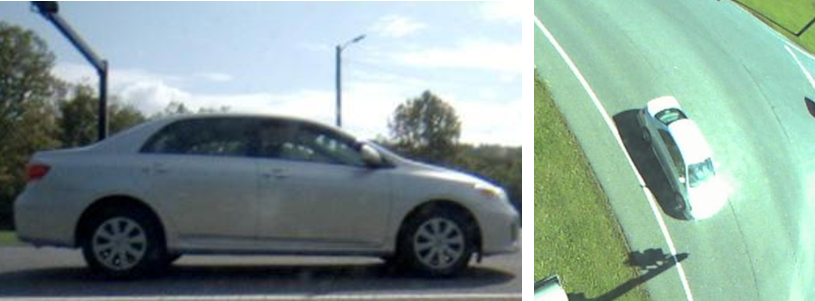}
\caption{Example images from the OOVD dataset (cropped). Left: side view of a Toyota Corolla, Right: Corresponding overhead view.}
\end{figure}

\begin{figure}[h]
\centering
\includegraphics[width=0.9\columnwidth]{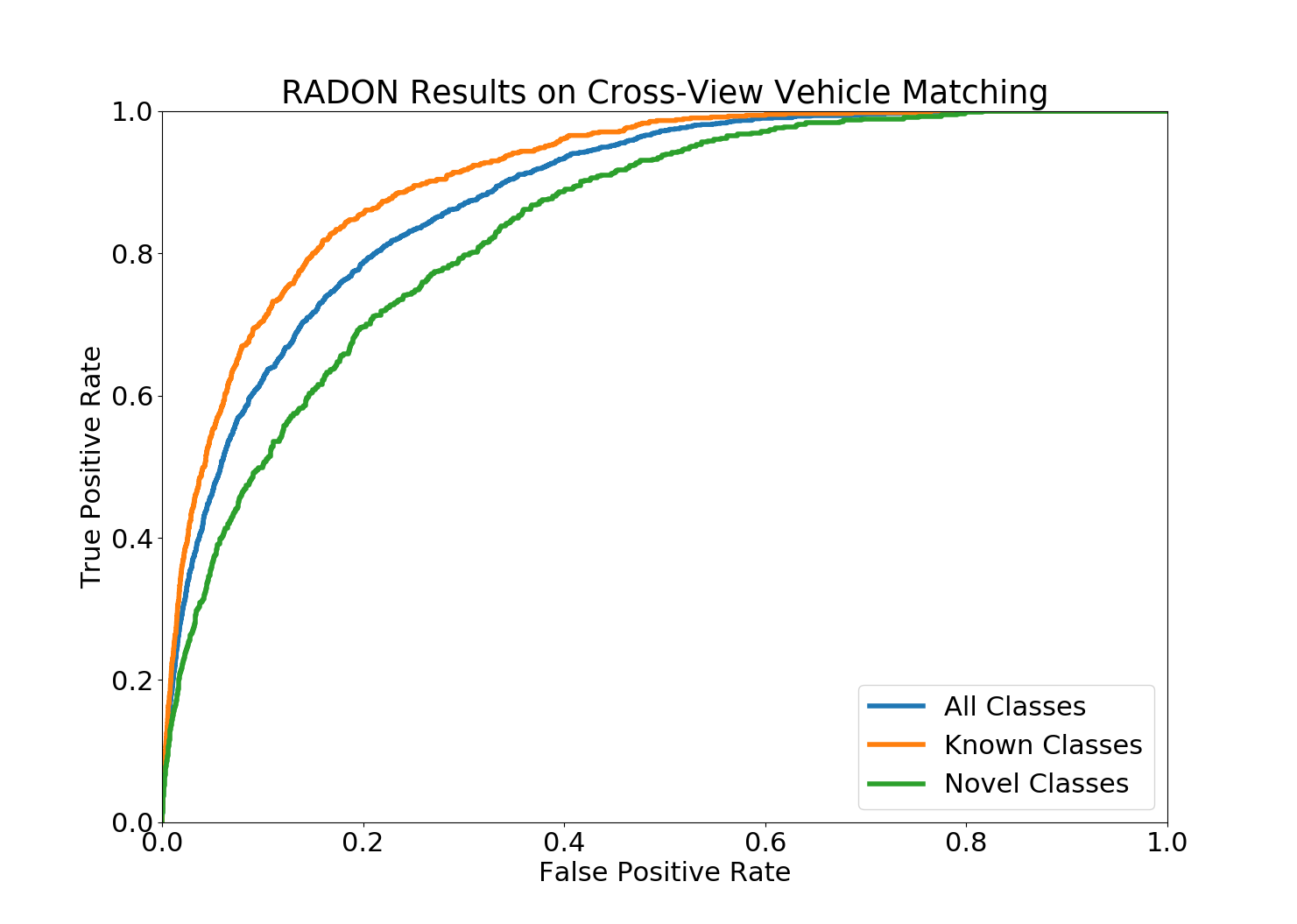}
\caption{ROC curve for RADON on OOVD. The network performs best on new instances of classes it trained on, but still retains much of its performance on previously unseen classes.}
\label{fig:roc_oovd}
\end{figure}

RADON performs well in this application, and most importantly it maintains this performance for new classes unseen in the training data. Figure~\ref{fig:roc_oovd} shows a receiver operating characteristic (ROC) curve for RADON on the OOVD dataset in three scenarios: known classes, or classes seen during training, compared with other known classes; novel classes, or classes not seen during training, compared with other novel classes; and all available classes. Unsurprisingly, the network performs best when comparing two classes that were both seen during training. However, it retains much of its performance even when evaluating two completely novel classes. This allows RADON to be applied beyond the scope of its original training data, providing a much more powerful tool for image processing. Generalization is particularly important in applications like the OOVD dataset. We cannot reasonably expect to build a training set with every vehicle type that a traffic camera would see, and if we did it would quickly become outdated. A network that handles novel classes dramatically reduces the amount of training data needed to handle the large space of vehicle classes. 

\begin{figure}[h]
\centering
\includegraphics[width=0.9\columnwidth]{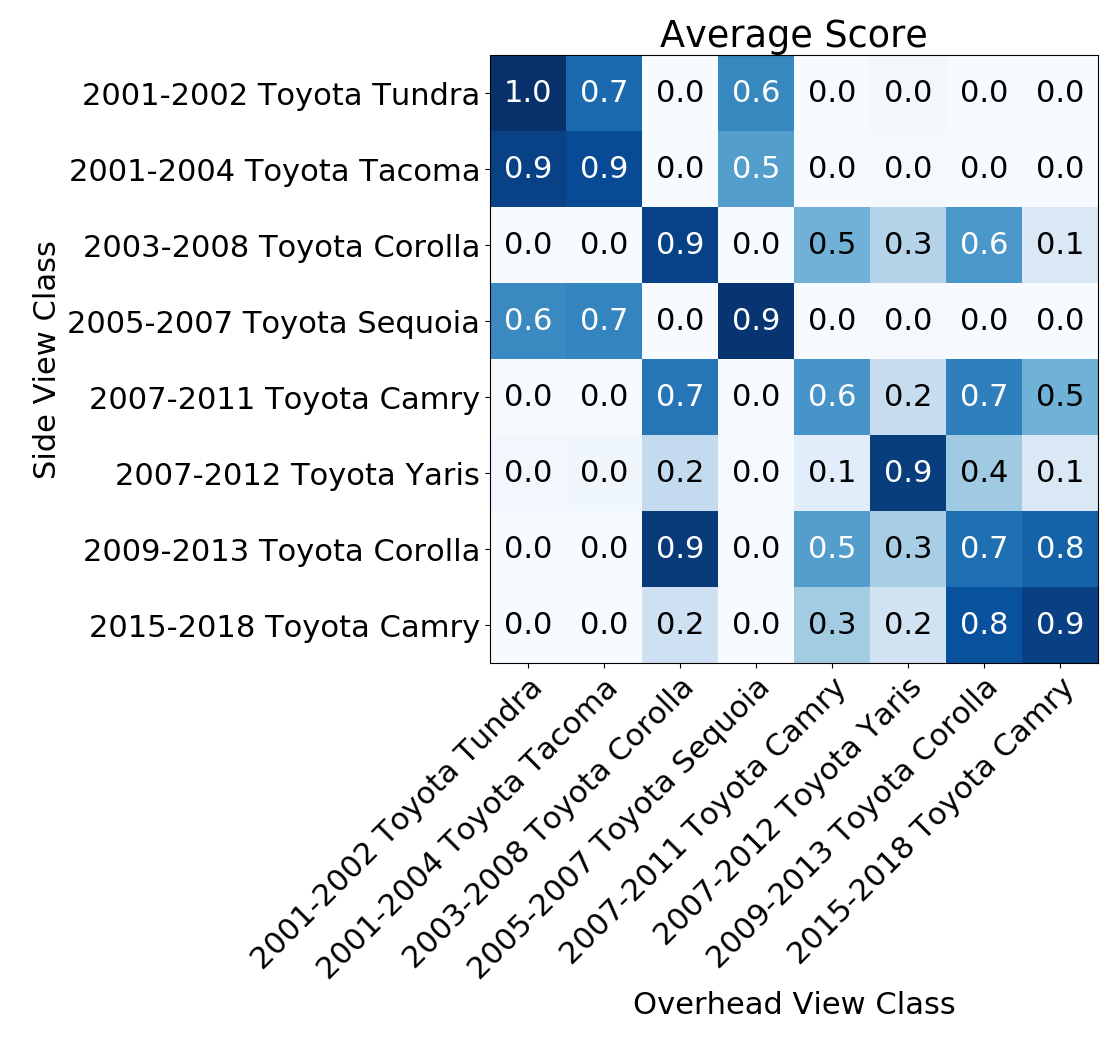}
\caption{Average score given by the network for a subset of classes. Each grid cell represents the average score assigned by the network given a side image from the vertical axis and an overhead image from the horizontal axis. A perfect network would result in 1.0 (dark blue) along the diagonal and zeros everywhere else.}
\label{fig:toyota_matrix}
\end{figure}

In addition to the overall performance, we consider the class-specific performance. The classes in OOVD are extremely specific, narrowing by both a specific make and model and range of years. Figure~\ref{fig:toyota_matrix} shows the model results on a subset of classes using Toyota vehicles. While the network generally performs well, the causes of confusion are also of interest. It struggles, for instance, at distinguishing two types of Toyota truck, the Tundra and Tacoma, and at distinguishing between two similar sedans, the Corolla and Camry. These errors reinforce that the network has learned useful features for distinguishing vehicles that accord with our intuition, and is confusing classes that are genuinely similar. 

The results in Figure~\ref{fig:toyota_matrix} show the network's performance for highly similar classes with small amounts of training data available. The number of training examples, ranging from just one up to 43 for a class, are shown in Table~\ref{tab:toyota_data}. The examples include variations in vehicle color, lighting conditions, and background, along with the two drastically different viewing angles. Given the challenges of this dataset, the results are encouraging and the sources of confusion reasonable.  
\begin{table}
\centering
\begin{tabular}{|c|c|}
\hline 
Class & Training Examples \\ 
\hline 
2001-2002 Toyota Tundra & 4 \\ 
\hline 
2001-2004 Toyota Tacoma & 13 \\ 
\hline 
2003-2008 Toyota Corolla & 2 \\ 
\hline 
2005-2007 Toyota Sequoia  & 1 \\ 
\hline 
2007-2011 Toyota Camry & 43 \\ 
\hline 
2007-2012 Toyota Yaris & 3 \\ 
\hline 
2009-2013 Toyota Corolla & 19 \\ 
\hline 
2015-2018 Toyota Camry  & 4 \\ 
\hline 
\end{tabular} 
\caption{Counts of training examples for Toyota classes}
\label{tab:toyota_data}
\end{table}

While not comprehensive, these results are promising. The network is able to re-identify targets across extremely different camera view angles, and to accomplish that distinction at a reasonable level of class granularity.



\section{Application to Person Identification}
The network was also applied to a person re-identification problem, in which we attempt to match facial images across multiple sensor types. In this case, images are captured from similar angles, and the network matches targets between RGB and infrared (IR) images. This application is particularly useful in subject tracking across multiple cameras, in which we need to recognize the same person in multiple modalities or form multiple angles. The SYSU-MM01 dataset consists of labeled RGB and IR images of 491 unique subjects \cite{rgbir}. We trained RADON using images from 441 subjects, with the remaining 50 identities held out for no-shot testing. For this application we crop out the top portion of the image, surrounding only the subject's face, and resize to the appropriate network input size. 

\begin{figure}
\centering
\includegraphics[height=0.5\columnwidth]{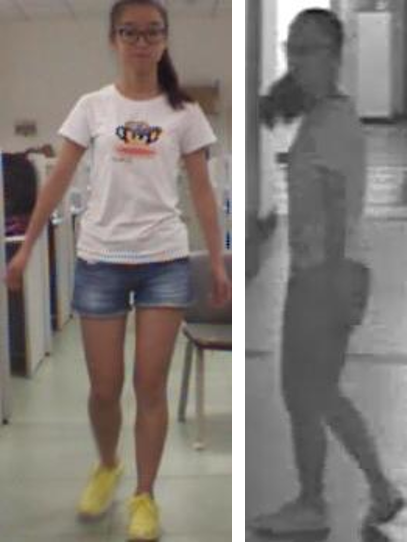}
\caption{Examples of SYSU data: A test subject seen from both RGB and IR cameras at different angles. The SYSU dataset includes a wide range of angles for each subject and both indoor and outdoor shots.}
\label{fig:sysu_data}
\end{figure}

The network performs reasonably well identifying matches between new instances of previously observed identities, and in distinguishing between a known identity and novel identities not seen in training. It struggles when both input images are previously unseen identities. In its current state, the network is thus most applicable for re-locating a particular subject of interest rather than identifying unknown subjects. 

\begin{figure}
\centering
\includegraphics[width=0.9\columnwidth]{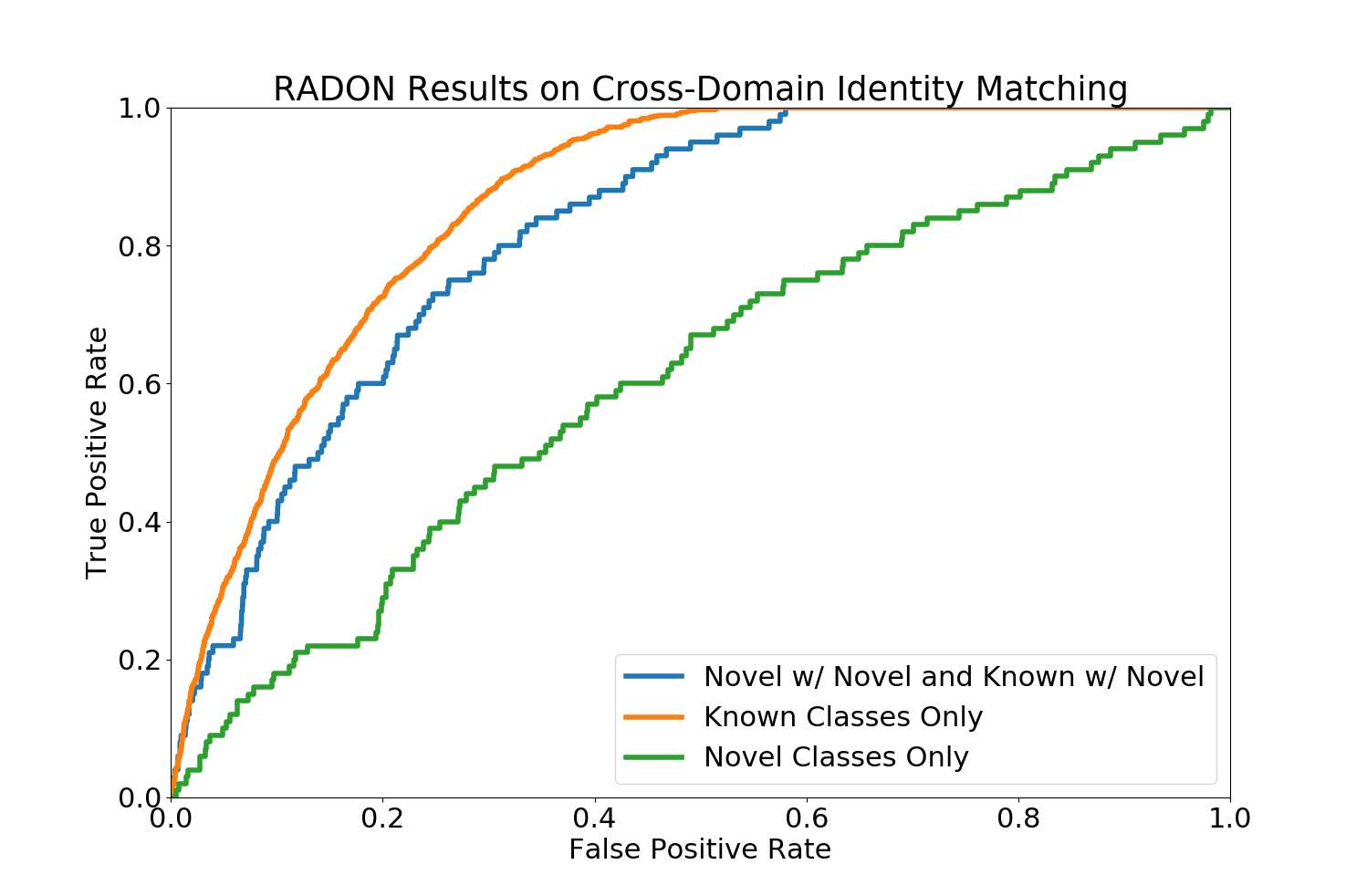}
\caption{RADON overall ROC Curve on RGB to IR person identity matching. The network struggles to distinguish between two new identities, but performs well with one new and one known identity.}
\label{fig:sysu_results}
\end{figure}

Additional training may improve the network's performance on novel identities. We also note that the SYSU data includes all angles, including angles from which the face is not visible. A more limited facial recognition dataset may result in better performance for novel identities, as more facial information is available. 


\section{Conclusion} 
We have presented RADON, a modified Siamese network architecture along with weights pre-trained on a different image processing task (RGB image classification) to perform matching across different domains, including no-shot matching. RADON performs well on the cross-view matching of vehicles, exhibiting understandable confusion in certain similar, unseen classes. The approach also shows strong performance on RGB-IR person identification, matching and rejecting non-matches even when novel classes are introduced. RADON's ability to extend to classes not found in the original dataset enables cross-domain matching in applications with numerous classes. 
We found that untying the weights in each branch of the Siamese architecture during training allowed the network to specialize to its input domain.
While work remains to fine-tune the architecture and parameters for different applications, the basic approach shows promise in re-identifying targets across image domains.

Future work will extend RADON to additional cross-domain problems and fine-tune the architecture and parameters for the current applications. 

\bibliographystyle{ieeetran}
\bibliography{bibliography}

\vspace{12pt}
\end{document}